\documentclass{article}

\usepackage{times}
\usepackage{graphicx} % more modern
\usepackage{subfigure} 
\graphicspath{ {images/} }

\usepackage{natbib}

\usepackage{algorithm}
\usepackage{algorithmic}

\usepackage{amsmath}

\usepackage{hyperref}

\usepackage[accepted]{icml2016}

\icmltitlerunning{Real-Time Anomaly Detection }

\begin{document} 

\twocolumn[
\icmltitle{Real-Time Anomaly Detection for Streaming Analytics}

\icmlauthor{Subutai Ahmad}{sahmad@numenta.com}
\icmladdress{Numenta, Inc.,
            791 Middlefield Road, Redwood City, CA 94063 USA}
\icmlauthor{Scott Purdy}{spurdy@numenta.com}
\icmladdress{Numenta, Inc.,
            791 Middlefield Road, Redwood City, CA 94063 USA}

\icmlkeywords{anomaly detection, streaming analytics, real-time analytics, HTM}

\vskip 0.3in
]

\begin{abstract} 
Much of the world’s data is streaming, time-series data, where anomalies give
significant information in critical situations. Yet detecting anomalies in
streaming data is a difficult task, requiring detectors to process data in
real-time, and learn while simultaneously making predictions. We present a novel
anomaly detection technique based on an on-line sequence memory algorithm called
Hierarchical Temporal Memory (HTM). We show results from a live
application that detects anomalies in financial metrics in real-time. We also
test the algorithm on NAB, a published benchmark for real-time anomaly
detection, where our algorithm achieves best-in-class results.
\end{abstract} 

\section{Introduction}
\label{introduction}

Across every industry, we are seeing an exponential increase in
the availability of streaming, time-series data. Largely driven by the rise of
the Internet of Things (IoT) and connected real-time data sources, we now have
an enormous number of applications with sensors that produce important,
continuously changing data.

The detection of anomalies in real-time streaming data has practical and
significant applications across many industries.  There are numerous use cases
for anomaly detection, including preventative maintenance, fraud prevention,
fault detection, and monitoring. The use cases can be found throughout numerous
industries such as finance, IT, security, medical, energy, e-commerce, and
social media.

\begin{figure}[ht]
\vskip 0.2in
\begin{center}
\centerline{\includegraphics[width=\columnwidth]{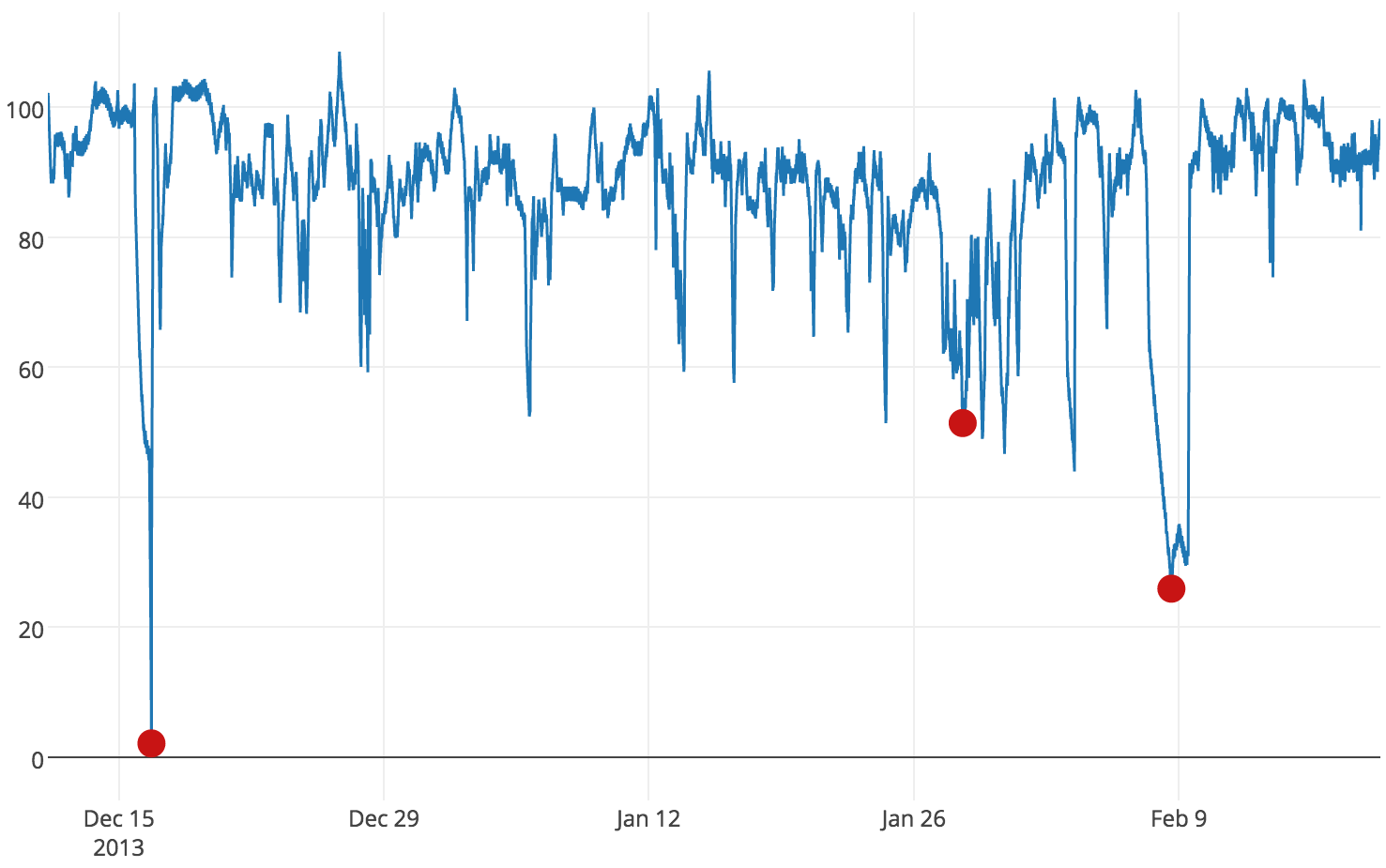}}
\caption{The figure shows real-world temperature sensor data from an internal
component of a large industrial machine. Anomalies are labeled with red circles.
The first anomaly was a planned shutdown. The third anomaly is a catastrophic
system failure. The second anomaly, a subtle but observable change in the
behavior, indicated the actual onset of the problem that led to the eventual
system failure.}
\label{icml-historical}
\end{center}
\vskip -0.2in
\end{figure}

We define an anomaly as a point in time where the behavior of the system is
unusual and significantly different from past behavior. Under this definition,
an anomaly does not necessarily imply a problem. A change might be for a
negative reason like the temperature sensor on an engine going up, indicating a
possible imminent failure. Or the change might be for a positive reason like web
clicks on a new product page are abnormally high, showing strong demand. Either
way, the data is unusual and may require action. Anomalies can be spatial,
meaning the value is outside the typical range like the first and third
anomalies in Figure \ref{icml-historical}. They can also be temporal, where
the value isn't outside the typical range but the sequence in which it occurs
is unusual. The middle anomaly in Figure \ref{icml-historical} is a temporal
anomaly.

Real-time applications impose their own unique constraints for machine
learning. Anomaly detection in streaming applications is particularly
challenging. The detector must process data and output a decision in real-time,
rather than making many passes through batches of files. In most scenarios the
number of sensor streams is large and there is little opportunity for human, let
alone expert, intervention. As such, operating in an unsupervised, automated
fashion (e.g. without manual parameter tweaking) is often a necessity. The
underlying system is often non-stationary, and detectors must continuously
learn and adapt to changing statistics while simultaneously making predictions.

The goal of this paper is to introduce a novel anomaly detection technique
designed for such real-time applications. We show how to use Hierarchical
Temporal Memory (HTM) networks \cite{Hawkins2016, Padilla2013, Rozado2012} in a
principled way to robustly detect anomalies in a variety of conditions. The
resulting system is efficient, extremely tolerant to noisy data, continously
adapts to changes in the statistics of the data, and detects very subtle
anomalies while minimizing false positives. We show qualitative examples from a
real-time financial anomaly detection application. We also report leading
results on an open benchmark for real-time anomaly detection. The algorithm has
been deployed commercially, and we discuss some of the practical lessons learned
from these deployments. We have made the complete source code (including end
application code) available in open source repositories
\footnote{Please see \url{http://numenta.org}}.

\section{Related Work}

\begin{figure}[ht]
\vskip 0.2in
\begin{center}
\centerline{\includegraphics[width=\columnwidth]{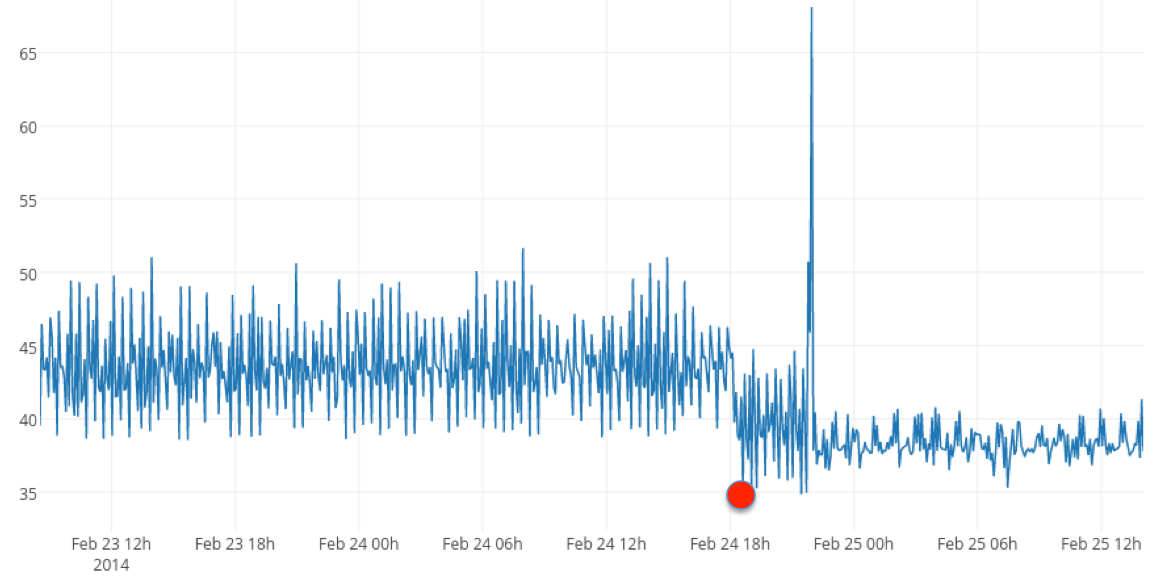}}
\caption{CPU utilization (percent) for an Amazon EC2 instance. A change to the
software running on the machine caused the CPU usage to change. Continuous
learning is essential for performing anomaly detection on streaming data like
this.}
\label{continuous-learning}
\end{center}
\vskip -0.2in
\end{figure}

Anomaly detection in time-series is a heavily studied area, dating back to
\cite{Fox1972}. Some techniques, like classification-based methods, are
supervised or semi-supervised. While labelled data can be used to improve
results, supervised techniques are typically unsuitable for anomaly detection
\cite{Gornitz2013}. Figure \ref{continuous-learning} illustrates the need for
continuous learning, which is not typically possible with supervised
algorithms.

Other techniques, like simple thresholds, clustering, and exponential smoothing,
are only capable of detecting spatial anomalies. Holt-Winters is an example of
the latter that is commonly implemented for commercial applications
\cite{Szmit2012}. Also commonly used in practice are change point detection
methods, which are capable of identifying temporal anomalies. The typical
approach is to model the time series in two independent moving windows and
detect when there is a significant deviation in the time series metrics
\cite{Basseville1993}. These methods are often extremely fast to compute and
have low memory overhead. The detection performance of these statistical
techniques can be sensitive to the size of the windows and thresholds. This
sometimes results in many false positives as the data changes, requiring
frequent updates to the thresholds in order to detect anomalies while minimizing
false positives.

The Skyline project provides an open source implementation of a number of
statistical techniques for detecting anomalies in streaming data
\cite{Stanway2013}. The different algorithms can be combined into an ensemble.
The Skyline algorithms are included in our results.

There are other algorithms capable of detecting temporal anomalies in complex
scenarios. ARIMA is a general purpose technique for modeling temporal data with
seasonality \cite{Bianco2001}. It is effective at detecting anomalies in data
with regular daily or weekly patterns. It is not capable of dynamically
determining the period of seasonality, although extensions have been developed
for doing so \cite{Hyndman2008}. A technique for applying ARIMA to multivariate
data has also been studied \cite{Tsay2000}. Bayesian change point detection
methods are a natural approach to segmenting time series and can be used for
online anomaly detection \cite{Adams2007,Tartakovsky2013}. Some
additional techniques for general purpose anomaly detection on streaming data
include \cite{Keogh2005, Rebbapragada2009}.

Yahoo released the open source EGADS framework for time series anomaly
detection that pairs time series forecasting techniques with common anomaly detection
algorithms \cite{laptev2015generic}. Twitter released its own open source anomaly
detection algorithms for time series data \cite{Kejariwal2015}. Both are
capable of detecting spatial and temporal anomalies. Empirical comparison with
Twitter's detection software are included in our results.

There have been a number of model-based approaches applied to specific domains.
These tend to be extremely specific to the domain they are modeling. Examples
include anomaly detection in aircraft engine measurements \cite{Simon2015},
cloud datacenter temperatures \cite{Lee2013}, and ATM fraud detection
\cite{Klerx2014}. While these approaches may have success in a specific domain,
they are not suitable for general purpose applications.

We have reviewed some of the algorithms most relevant to our work. A
comprehensive literature review is outside the scope of this paper but there are
several thorough reviews of anomaly detection techniques for further reading
\cite{Chandola2009, Hodge2004, Chandola2008}.

In this paper we focus on using Hierarchical Temporal Memory (HTM) for anomaly
detection. HTM is a machine learning algorithm derived from neuroscience
that models spatial and temporal patterns in streaming data \cite{Hawkins2016,
Rozado2012}. HTM compares favorably with some state of the art algorithms in
sequence prediction, particularly complex non-Markovian sequences \cite{Cui2015,
Padilla2013}. HTMs are continuously learning systems that automatically adapt
to changing statistics, a property particularly relevant to streaming analytics.

\section{Anomaly Detection Using HTM}

Typical streaming applications involve analyzing a continuous stream of data
occurring in real-time. Such applications contain some unique challenges. We
formalize this as follows. Let the vector $\boldsymbol{x}_t$ represent the state
of a real-time system at time $t$. The model receives a continuous stream of
inputs:

\begin{equation}
\ldots,\boldsymbol{x}_{t-2},\boldsymbol{x}_{t-1},
\boldsymbol{x}_t,\boldsymbol{x}_{t+1},\boldsymbol{x}_{t+2},\ldots
\end{equation}

Consider for example, the task of monitoring a datacenter. Components of
$\boldsymbol{x}_t$ might include CPU usage for various servers, bandwidth
measurements, latencies of servicing requests, etc.  At each point in time $t$ we would
like to determine whether the behavior of the system up to that point is
unusual.  One of the key challenges is that the determination must be made in
real-time, i.e. before time $t+1$ and without any look ahead. In practical
applications, the statistics of the system can change dynamically. For example,
in a production datacenter, software upgrades might be installed at any time that
alter the behavior of the system (Figure~\ref{continuous-learning}). Any
retraining of a model must be done
on-line, again before time $t+1$.  Finally, the individual measurements are
not independent and contain significant temporal patterns that can be
exploited.

HTM is a learning algorithm that appears to match the
above constraints. HTM networks are continuously learning and model the
spatiotemporal characteristics of their inputs.  HTMs have been shown to work well
for prediction tasks \cite{Cui2015, Padilla2013} but HTM networks do not directly output an
anomaly score. In order to perform anomaly detection we utilize two
different internal representations available in the HTM. Given an input
$\boldsymbol{x}_t$, the vector $\mathbf{a}(\boldsymbol{x}_t)$ is a sparse binary
code representing the current input. We also utilize an internal state vector
$\boldsymbol{\pi}(\boldsymbol{x}_t)$ which represents a {\em prediction} for
$\mathbf{a}(\boldsymbol{x}_{t+1})$, i.e. a prediction of the next input
$\boldsymbol{x}_{t+1}$. The prediction vector incorporates inferred information
about current sequences. In particular, a given input will lead to different
predictions depending on the current detected sequence and the current inferred
position of the input within the sequence.  The quality of the prediction is
dependent on how well the HTM is modeling the current data stream. See
\cite{Hawkins2016} for a more detailed explanation of these representations.

\begin{figure}[ht]
\vskip 0.2in
\begin{center}
\centerline{\includegraphics[width=\columnwidth]{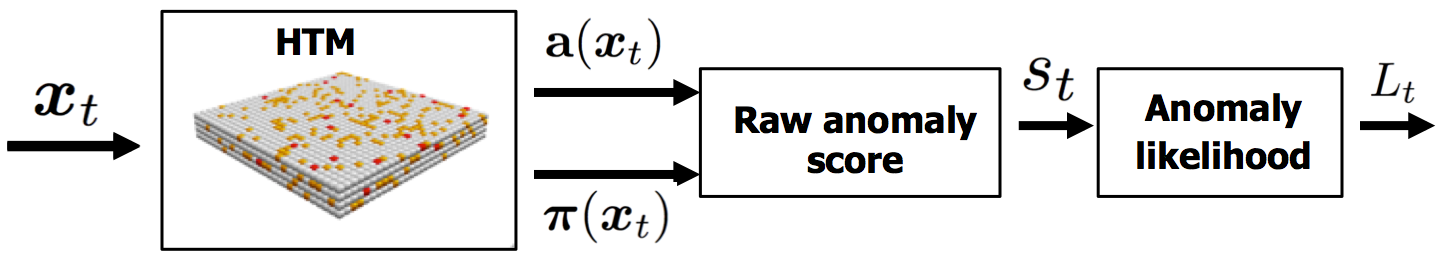}}
\caption{The primary functional steps in our algorithm.}
\label{block-diagram}
\end{center}
\vskip -0.2in
\end{figure}

$\mathbf{a}(\boldsymbol{x}_t)$ and $\boldsymbol{\pi}(\boldsymbol{x}_t)$ are
recomputed at every iteration but do not directly represent anomalies. In order
to create a robust anomaly detection system we introduce two additional steps.
We first compute a {\em raw anomaly score} from the two sparse vectors. We then
compute an {\em anomaly likelihood} value which is thresholded to determine
whether the system is anomolous. Figure~\ref{block-diagram} shows a block
diagram of our algorithm.  These two steps are detailed below.  We then describe
how to robustly handle a larger system consisting of multiple distinct models.

\subsection{Computing the Raw Anomaly Score}
We compute a raw anomaly score that measures the deviation between the model's
predicted input and the actual input. It is computed from the
intersection between the predicted and actual sparse vectors. At time $t$ the
raw anomaly score, $s_t$, is given as:

\begin{equation}
s_t = 1 - \frac{\boldsymbol{\pi}(\boldsymbol{x}_{t-1})
    \cdot \mathbf{a}(\boldsymbol{x}_t)}{|\mathbf{a}(\boldsymbol{x}_t)|}
\end{equation}

%\begin{equation}
%s_t = 1 - \frac{\boldsymbol{\pi}(\boldsymbol{x}_{t-1})
%\bigcap \mathbf{a}(\boldsymbol{x}_t)}{|\mathbf{a}(\boldsymbol{x}_t)|}
%    = 1 - \frac{\boldsymbol{\pi}(\boldsymbol{x}_{t-1})
%    \cdot \mathbf{a}(\boldsymbol{x}_t)}{|\mathbf{a}(\boldsymbol{x}_t)|}
%\end{equation}
%
The raw anomaly score will be $0$ if the current input is perfectly predicted,
$1$ if it is completely unpredicted, or somewhere in between depending on the
similarity between the input and the prediction.

An interesting aspect of this score is that branching sequences are handled
correctly. In HTMs, multiple predictions are represented in
$\boldsymbol{\pi}(\boldsymbol{x}_t)$ as a binary union of each individual
prediction. Similar to Bloom filters, as long as the vectors are sufficiently
sparse and of sufficient dimensionality, a moderate number of predictions can be
represented simultaneously with exponentially small chance of error
\cite{Bloom1970,Ahmad2016}. The anomaly score handles branching sequences
gracefully in the following sense. If two completely different inputs are both
possible and predicted, receiving either input will lead to a $0$ anomaly score.
Any other input will generate a positive anomaly score.

Changes to the underlying system are also handled gracefully due to the
continuous learning nature of HTMs. If there is a shift in the behavior of the
system, the anomaly score will be high at the point of the shift, but will
automatically degrade to zero as the model adapts to the ``new normal''. Shifts
in the temporal characteristics of the system are handled in addition to
spatial shifts in the underlying metric values. (See Results section for
some examples.)

\begin{figure}[ht]
\vskip 0.2in
\begin{center}
\centerline{\includegraphics[width=\columnwidth]{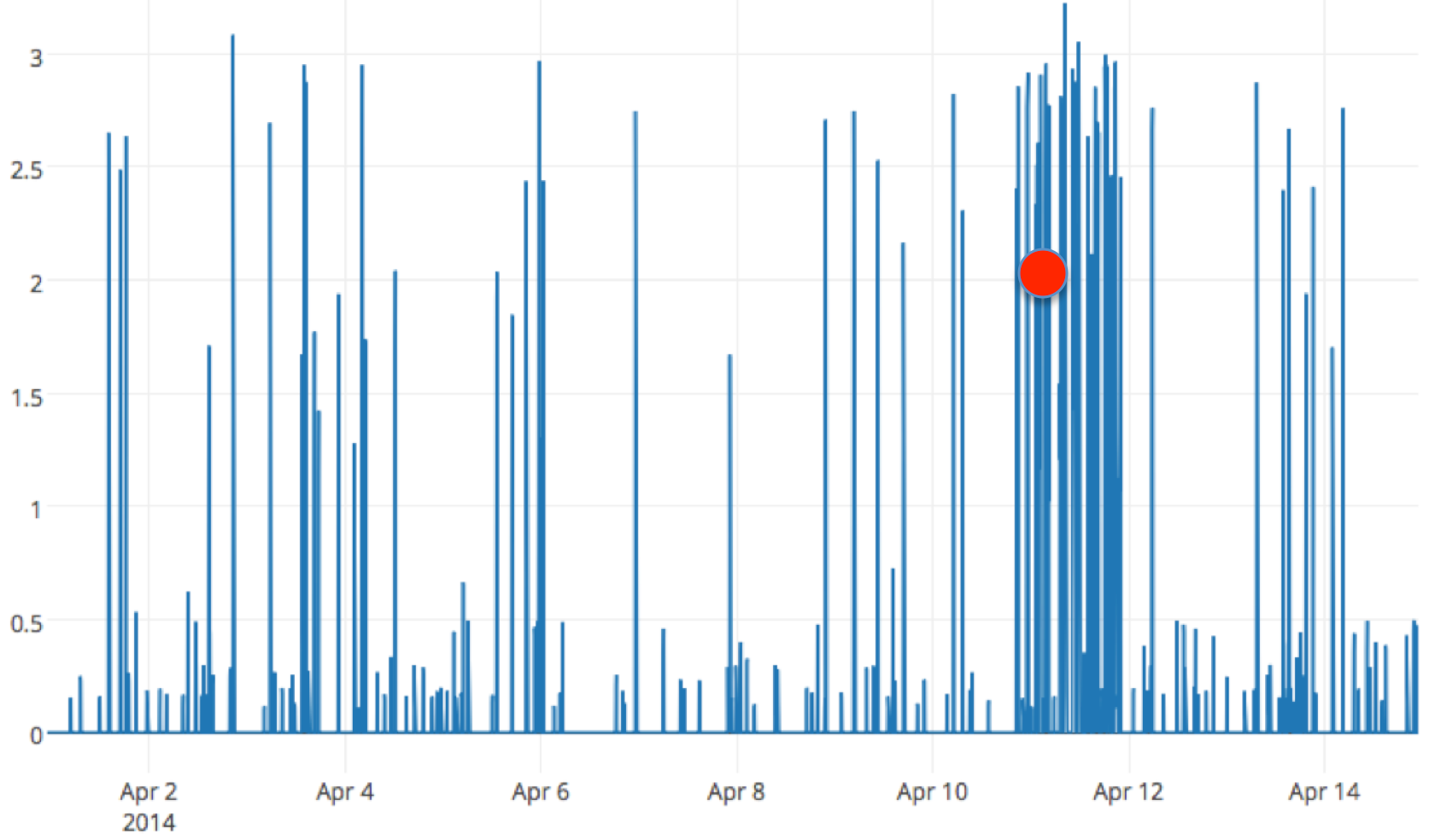}}
\caption{A very noisy, unpredictable stream. The data shows the
latency (in seconds) of a load balancer on a production website. The red dot shows the
approximate location of an unusual increase in latencies.}
\label{lb}
\end{center}
\vskip -0.2in
\end{figure}

\subsection{Computing Anomaly Likelihood}

The raw anomaly score described above represents an instantaneous measure of the
predictability of the current input stream. This works well for
predictable scenarios but in many practical applications, the underlying system
is inherently noisy and unpredictable. In these situations it is often the
change in predictability that is indicative of anamolous behavior. As an
example, consider Figure~\ref{lb}. This data shows the latency of a load balancer in
serving HTTP requests on a production web site.  Although the latency is
generally low, it is not unusual to have occasional random jumps, and the
corresponding spike in anomaly score. Thresholding the raw anomaly score directly
would lead to many false positives. However, a sustained increase in the
frequency of high latency requests, as shown in the second half of the figure,
is unusual and thus reported as an anomaly.

To handle this class of scenarios, we introduce a second step. Rather than
thresholding the raw score directly, we model the distribution of anomaly scores
and use this distribution to check for the likelihood that the current state is
anomalous. The anomaly likelihood is thus a metric defining how anomalous the
current state is based on the prediction history of the HTM model. To compute
the anomaly likelihood we maintain a window of the last $W$ raw anomaly scores.
We model the distribution as a rolling normal distribution where the sample mean
and variance are continuously updated from previous anomaly scores as follows:

\begin{equation}
\label{eq:mut}
\mu_t = \frac{\sum_{i=0}^{i=W-1}s_{t-i}}{k}
\end{equation}

\begin{equation}
\sigma_t^2 = \frac{\sum_{i=0}^{i=W-1} (s_{t-i} - \mu_t)^2}{k-1}
\end{equation}

We then compute a recent short term average of anomaly scores, and apply a
threshold to the Gaussian tail probability (Q-function, \cite{Karagiannidis2007}) to decide
whether or not to declare an anomaly\footnote{The Guassian tail probability is
the probability that a Gaussian variable will obtain a value larger than $x$
standard deviations above the mean.}. We define the {\em anomaly likelihood} as
the complement of the tail probability:

\begin{equation}
\label{eq:lt}
L_t = 1 - Q\left(\frac{\tilde{\mu}_t-\mu_t}{\sigma_t}\right)
\end{equation}

where:

\begin{equation}
\tilde{\mu}_t = \frac{\sum_{i=0}^{i={W'}-1}s_{t-i}}{j}
\end{equation}

$W'$ here is a window for a short term moving average, where $W' \ll W$.
We threshold $L_t$ and report an anomaly if it is very close to $1$:

\begin{equation}
\label{eq:ltest}
\text{anomaly detected} \equiv L_t \geq 1 - \epsilon
\end{equation}

It is important to note that this test is applied to the distribution of anomaly
scores, not to the distribution of underlying metric values
$\boldsymbol{x}_{t}$. As such, it is a measure of how well the model is able to
predict, relative to the recent history. In clean predictable scenarios $L_t$ behaves
similarly to $s_t$. In these cases the distribution of scores will have very
small variance and will be centered near $0$. Any spike in $s_t$ will similarly
lead to a corresponding spike in $L_t$. However in scenarios with some inherent
randomness or noise, the variance will be wider and the mean further from $0$. A single
spike in $s_t$ will not lead to a significant increase in $L_t$ but a series of
spikes will.  Interestingly, a scenario that goes from wildly random to
completely predictable will also trigger an anomaly.

Since thresholding $L_t$ involves thresholding a tail probability,
there is an inherent upper limit on the number of alerts. With $\epsilon$
very close to $0$ it would be unlikely to get
alerts with probability much higher than $\epsilon$. This also imposes an upper
bound on the number of false positives. Under the assumption that anomalies
themselves are also extremely rare, the hope is that the ratio of true positives
to false positives is always in a healthy range (see Results below).

Although we use HTM as the underlying temporal model, the likelihood technique is
not specific to HTMs. It could be used with any other algorithm that outputs a
sparse code or scalar anomaly score. The overall quality of the detector
will be dependent on the ability of the underlying model to represent the
domain.

\subsection{Combining Multiple Independent Models}

Many industrial or complex environments contain a large number of sensory data
streams. In theory, given sufficient resources, it is possible to create one
large complex model with the entire vector stream as input. In practice, it is
common to decompose a large system into a number of smaller models. It is easier
to train smaller models because the complexity of training and inference grows
much faster than linearly with the size of the input dimensionality
\cite{Bishop2006}. Thus decomposing the solution into a set of smaller models
may lead to improved accuracy and much faster performance. However, even with
such a decomposition, it is important to compute a global measure that
accumulates the results of individual models and indicates whether those
portions of the system are in an unusual state. As an example, consider a
datacenter running a production website. An automated alerting system might need
to continually decide whether to generate an alarm and possibly wake up the
on-call engineer.

\begin{figure}[ht]
\vskip 0.2in
\begin{center}
\centerline{\includegraphics[width=\columnwidth]{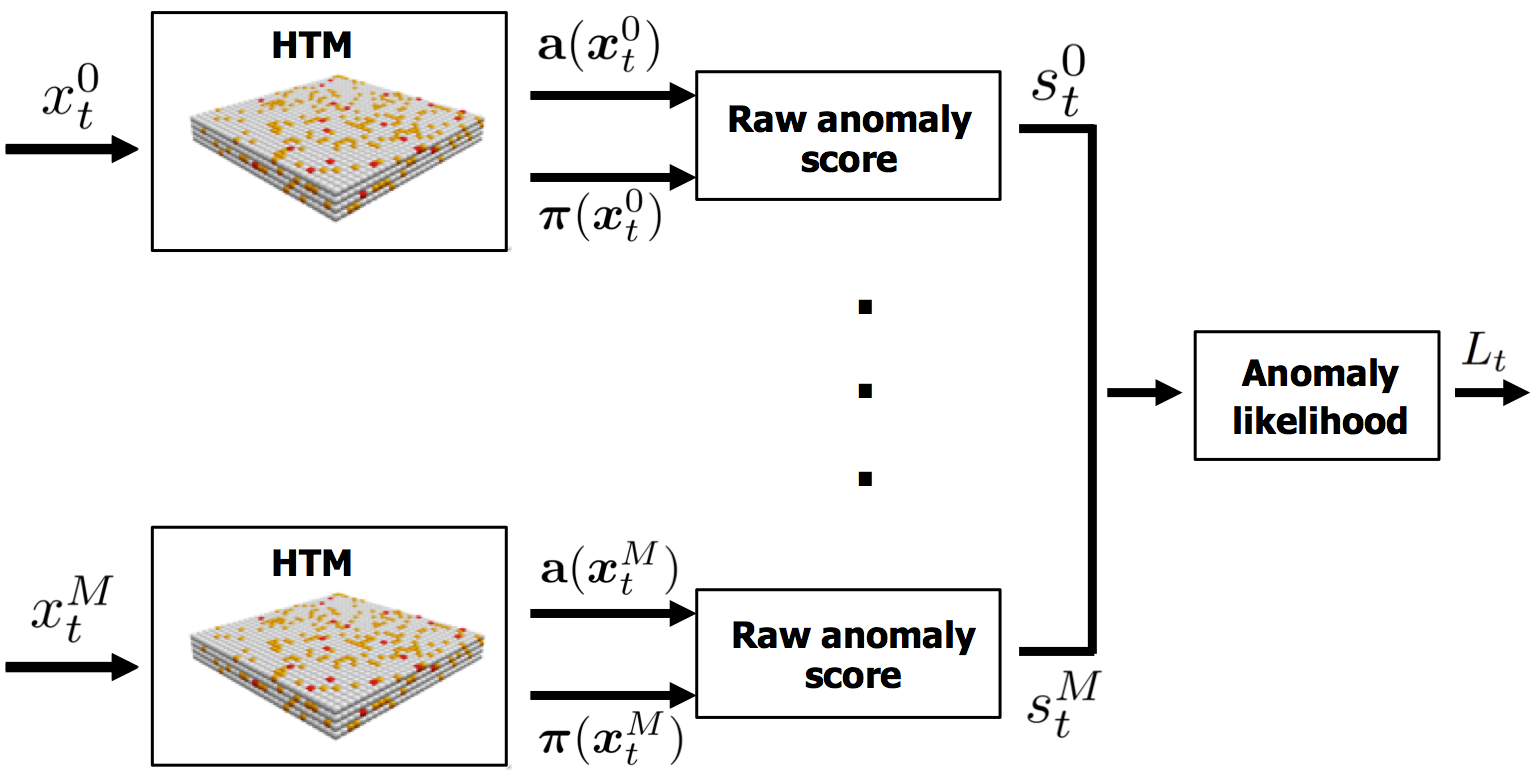}}
\caption{Functional diagram illustrating a complex system with multiple
independent models.}
\label{block-diagram-multi}
\end{center}
\vskip -0.2in
\end{figure}

We assume the inputs representing the system are broken up into $M$ distinct
models. Let $\boldsymbol{x}_t^m$ be the input at time $t$ to the $m$'th model,
and $s_t^m$ be the raw anomaly scores associated with each model. We wish to
compute a global metric indicating the overall likelihood of an anomaly in the
system (see Figure~\ref{block-diagram-multi}).

%Let $x_t^0$ $s_t^0$ $x_t^M$ $s_t^M$
%and $L_t^m$ be the raw anomaly scores and anomaly likelihoods
%associated with each model.
%output vector $\mathbf{a}(\boldsymbol{x}_t^0)$
%$\mathbf{a}(\boldsymbol{x}_t^M)$
%is a sparse
%binary vector representing the current input. The output vector
%$\boldsymbol{\pi}(\boldsymbol{x}_t^0)$
% $\boldsymbol{\pi}(\boldsymbol{x}_t^M)$
% is a {\em prediction} of the sparse
%representation for a future time step $\boldsymbol{x}_{t+1}$.

One possible
approach is to estimate the joint distribution $P(s_t^0,\ldots,s_t^{M-1})$ and
apply a threshold to the tail probability. It can be challenging
to model the joint distribution, particularly in a streaming context. If we
further assume the models are independent, we could simplify and instead estimate:

\begin{equation}
P(s_t^0,\ldots,s_t^{M-1}) = \prod_{i=0}^{i=M-1}P(s_t^i)
\end{equation}

Given this, a version of our anomaly likelihood can be computed as:

\begin{equation} \label{eq:multi1}
1 - \prod_{i=0}^{i=M-1}Q\left(\frac{\tilde{\mu}_t^i-\mu_t^i}{\sigma_t^i}\right)
\end{equation}

%\begin{figure*}[ht]
%\vskip 0.2in
%\begin{center}
%\centerline{\includegraphics[width=0.6\textwidth]{multi_metrics}}
%\caption{A diagram illustrating the likelihood for three different models
%in three scenarios. (b) represents a high likelihood of anomalies since
%all three metrics spike at the same time. Without temporal windowing, (a) and (c)
%would be given identical (low) weight. With temporal windowing, the likelihood
%of anomalies in (c) is higher than (a) but lower than (b).}
%\label{multi-metrics}
%\end{center}
%\vskip -0.2in
%\end{figure*}
%
There is one flaw with the above methodology. In real-time dynamic scenarios,
critical problems in one part of the system can often cascade to other areas.
Thus there are often random temporal delays built in, which can in turn lead to
different temporal delays between anomaly scores in the various models
\cite{Kim2013}. For example, a situation where multiple unusual events occur
close to one another in different models is far more unlikely and unusual than a
single event in a single model. It is precisely these situations which are
valuable to detect and capture in a complex system.

Ideally we would be able to estimate a joint distribution of anomaly scores that
go back in time, i.e. $P(s_{t-j}^0,s_{t-j}^1,\ldots,s_t^{M-2},s_t^{M-1})$. In
theory this would capture all the dependencies, but this is even harder to
estimate than the earlier joint probability. Alternatively, in situations where
the system's topology is relatively
clear and under your control, it is possible to create an explicit graph of
dependencies, monitor expected behavior between pairs of nodes, and detect
anomalies with respect to those expectations. This technique has been shown to
enable very precise determination of anomalies in websites where
specific calls between services are monitored \cite{Kim2013}. However in most
applications this technique is also impractical. It may be difficult to model
the various dependencies and it is often infeasible to instrument arbitrary
systems to create this graph.

We would like to have a system that is fast to compute, makes relatively few
assumptions, and is adaptive. We propose a simple general purpose mechanism for
handling multiple models by modifying Eq.~(\ref{eq:multi1}) to incorporate a
smooth temporal window. A windowing mechanism allows the system to incorporate
spikes in likelihood that are close in time but not exactly coincident.  Let $G$
be a Gaussian convolution kernel:

\begin{equation}
G(x;\sigma) = \frac{1}{\sqrt{2\pi}\sigma}\exp^{-\frac{x^2}{2\sigma^2}}
\end{equation}

We apply this convolution to each individual model to obtain a final anomaly
likelihood score\footnote{Since this is a real-time system with no look-ahead,
the kernel only applies to time $\leq t$. As such we use a half-normal distribution
as the kernel, hence the additional factor of 2.}:

\begin{equation} \label{eq:multi2}
L_t = 1 - \prod_{i=0}^{i=M-1}2(G \ast Q)(\frac{\tilde{\mu}_t^i-\mu_t^i}{\sigma_t^i})
\end{equation}

As before, we detect an anomaly if the combined anomaly likelihood is greater
than a threshold $L_t \geq 1 - \epsilon$. Eq.~\ref{eq:multi2} represents a
principled yet pragmatic approach for detecting anomalies in complex real time
streaming applications\footnote{Due to numerical precision issues with products
of probabilities, in our implementation we follow common practice and use
summation of log probabilities.}. As before, $L_t$ is an indirect measure,
computed on top of the raw anomaly scores from each model. It reflects the
underlying predictability of the models at a particular point in time and does
not directly model the sensor measurements themselves.

\subsection{Practical Considerations}

There are three parameters in the single model scenario: $W$, $W'$, and
$\epsilon$. $W$ is the duration for computing the distribution of anomaly
scores. The system performance is not sensitive to $W$ as long as it is large
enough to compute a reliable distribution. The number $W'$ controls the short
term average of anomaly scores. In all our experiments below, we use a generous
value of $W=8000$ and $W'=10$.

The parameter $\epsilon$ is perhaps the most important parameter. It controls
how often anomalies are reported, and the balance between false positives and
false negatives. In practice we have found that $\epsilon = 10^{-5}$ works well
across a large range of domains. Intuitively, this should represent one false
positive about once every 10,000 records.

The muliple model scenario introduces one additional parameter, the window width
$\sigma$.  This is somewhat domain dependent but due to the soft Gaussian
convolution, the system is not extremely sensitive to this setting. In all our
experiments below, $\sigma=6$.

Computational efficiency is important for real-time applications. In our
implementation each model requires less than 10 milliseconds per input vector on
a current high end laptop. A server based deployment running multiple models in
parallel can run about 5,000 models on a high end server. This figure assumes
that each input arrives once every 5 minutes (thus each of the 5,000 models
needs to output one score every 5 minutes). As a service to the community, we
have released the complete source code to the algorithms as well as the server
application codebase as open source\footnote{Please see
\url{https://github.com/numenta}, specifically the {\bf nupic} and {\bf numenta-apps}
repositories.}.

\begin{figure}[ht]
\vskip 0.1in
\begin{center}
\centerline{\includegraphics[width=\columnwidth]{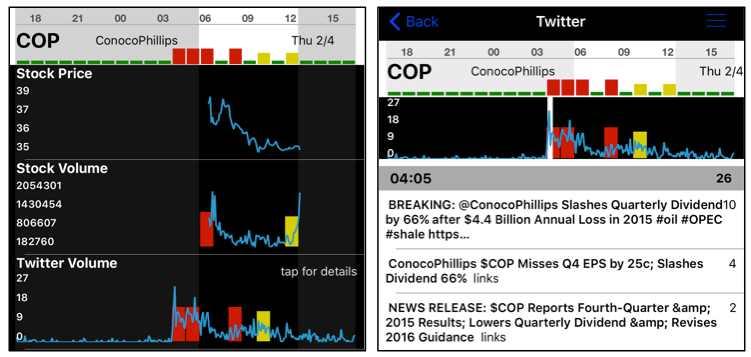}}
\caption{Real-time stock anomalies for Feb 4, 2016.}
\label{htmstocks}
\end{center}
\vskip -0.1in
\end{figure}

\section{Results}

We first show examples from a deployed application that
qualitatively demonstrates the behavior of our algorithm. We then show
quantitative results on benchmark data.

\begin{figure}[ht]
\vskip 0.1in
\begin{center}
\centerline{\includegraphics[width=\columnwidth]{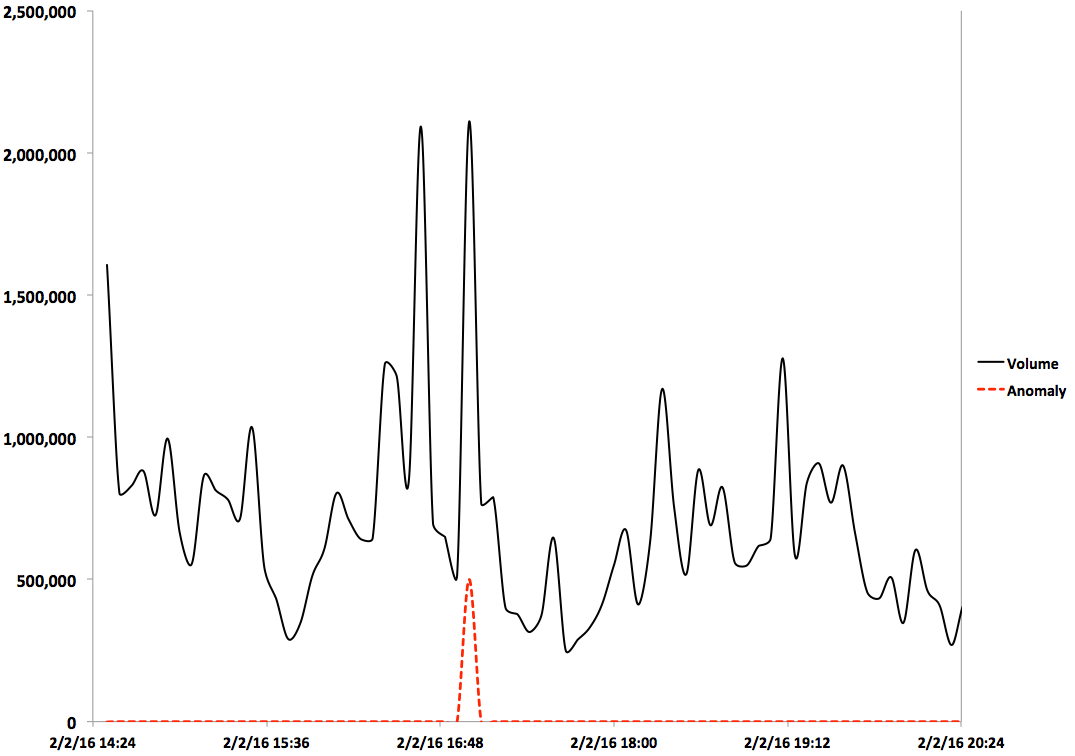}}
\caption{Stock trading volume for Facebook on Feb 2, 2016. The red dashed line
shows the point at which our algorithm detected an anomaly. In this stream
two consecutive spikes are highly unusual and represent an anomaly.}
\label{fb_volume}
\end{center}
\vskip -0.1in
\end{figure}

We have integrated our anomaly detection algorithm into a live real-time
product. The application continuously monitors a large set of financial and
social media metrics for a range of securities and alerts users in real time
when significant anomalies occur. Figure~\ref{htmstocks} shows two screenshots
from our application that demonstrate the value of real-time anomaly detection
to end users. In this example, an anomaly in the volume of Twitter activity
(related to a decrease in dividends) preceded a sharp drop in stock price.  The
Twitter anomaly occurred well before market opening. The underlying data streams
are extremely noisy and many of the important anomalies are temporal in nature.
Figure~\ref{fb_volume} shows raw data extracted from our application
demonstrating one such anomaly. The figure plots stock trading volume for
Facebook for a few hours. Each point represents average trading volume for a
period of five minutes. It is normal to see a single spike in trading volume but
it is extremely unusual to see two consecutive spikes. Two consecutive spikes therefore
represents a temporal anomaly in this stream. The red dashed line
indicates the point at which our algorithm detected the anomaly, i.e. the point
in time where $L_t \geq 1 - 10^{-5}$.

\begin{figure}[ht]
\vskip 0.1in
\begin{center}
\centerline{\includegraphics[width=\columnwidth]{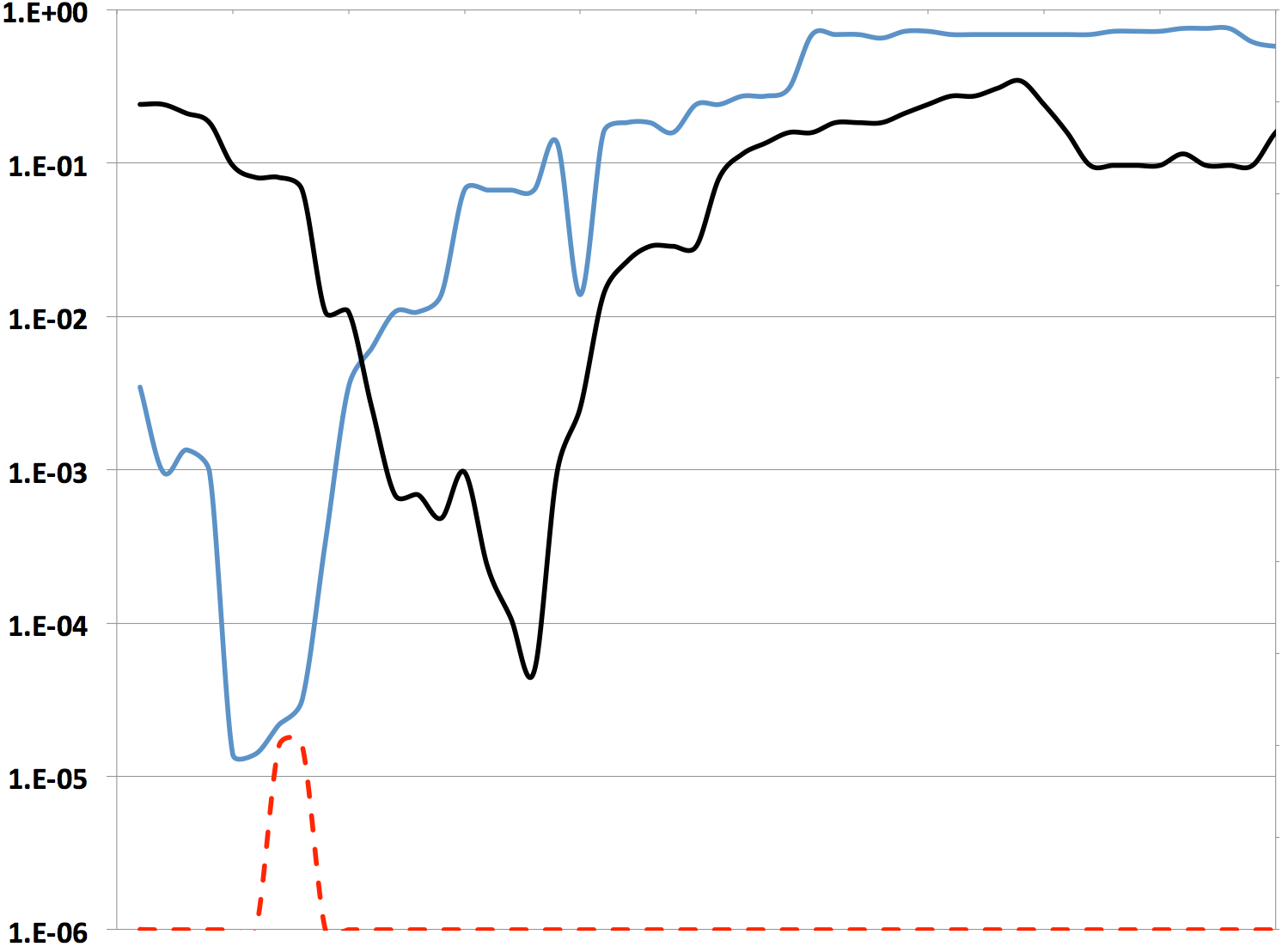}}
\caption{Q-values derived from two Comcast metrics on Feb 4, 2016. The red
dashed line shows the point at which our algorithm detected an anomaly.}
\label{multi_plot}
\end{center}
\vskip -0.1in
\end{figure}

Figure~\ref{multi_plot} shows a more detailed example demonstrating the value of
combining multiple metrics.  The two solid lines indicate the Q-values in
Eq~(\ref{eq:lt}) for two separate metrics related to Comcast shares. The lower
the curve, the more likely it is that the underlying metric is anomalous. When
looking at single models independently, an anomaly would only be detected if one
of these values drops below $\epsilon = 10^{-5}$. In this case no anomaly
would be detected since neither one dips below that value. However, as shown by
the drops, both metrics are behaving unusually. The red dashed line shows the
result of detecting an anomaly based on our combined metric,
Eq.~(\ref{eq:multi2}). Because the blue curve is so close to $10^{-5}$,
incorporating both metrics correctly flags an anomaly as soon as the black curve
dips below $10^{-1}$. As can be appreciated from the chart, it is difficult to
line up two metrics precisely, hence the value of the smooth temporal window.

\subsection{Results on Real-World Benchmark Data}

The above sections demonstrate qualitative results on individual data streams.
In this section we provide quantitative benchmark results. NAB is a benchmark
containing 58 streams with over 350,000 records of real-time streaming data
taken from a variety of different applications \cite{Lavin2015}. The dataset is
labeled with anomalies. The first 15\% of each data file is reserved for
auto-calibration. NAB also includes a window around each anomaly, and
incorporates a time-sensitive scoring mechanism that favors early detection (as
long as detections are within the anomaly windows). "Application profiles"
define the weighting for false positives and false negatives to illustrate
scenarios where fewer missed detections or fewer erroneous detections are more
valued. The benchmark requires models to use a single set of parameters across
all streams to mimic automation in real-world deployments.

\begin{table}[t]
\begin{center}
\begin{small}
\begin{sc}
\begin{tabular}{lcccr}
\hline
\abovespace\belowspace
Algorithm & NAB Score & Low FP & Low FN \\
\hline
\abovespace
Perfect          & 100.0 & 100.0   & 100.0 \\
HTM             & 65.3 & 58.6  & 69.4 \\
Twitter ADVec   & 47.1  & 33.6   & 53.5 \\
Etsy Skyline    & 35.7  & 27.1   & 44.5 \\
Bayes. Change Pt. & 17.7 & 3.2 & 32.2 \\
Sliding Threshold & 15.0 & 0.0 & 30.1 \\
\belowspace
Random          & 11.0  & 1.2    & 19.5 \\
\hline
\end{tabular}
\end{sc}
\end{small}
\end{center}
\vskip -0.1in
\label{results-table}
\caption{Performance on the NAB benchmark. The first column indicates
the standard NAB score for each algorithm. The last two columns indicate the
score for the ``prefer low FP'' and ``prefer low FN'' profiles of NAB,
representing two specific points on the ROC curve. For comparison we include
a ``Perfect'' detector, an idealized detector that makes no mistakes. }
\end{table}

\begin{figure*}[ht]
\vskip 0.2in
\begin{center}
\centerline{\includegraphics[width=\textwidth]{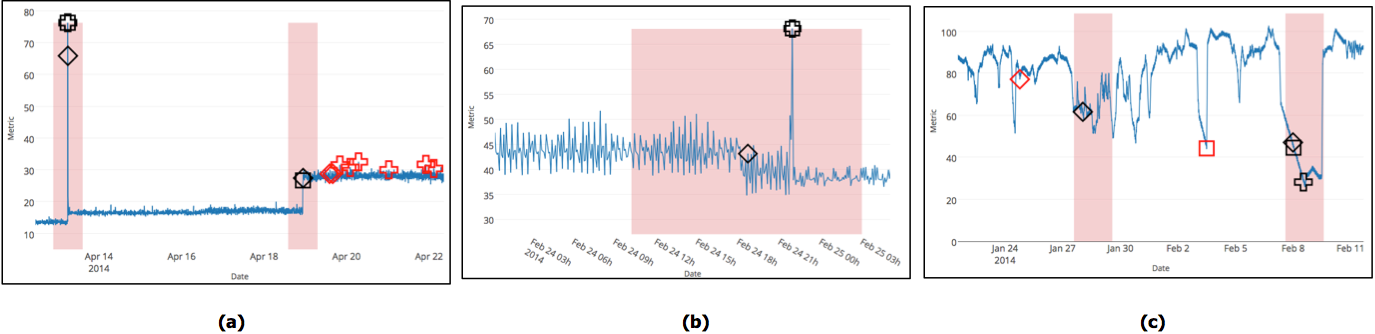}}
\caption{Example NAB results for three different data streams. The shapes
correspond to different detectors: HTM, Skyline, and ADVec are diamond, square,
and plus respectively. True positives are labeled in black, and false positives
are colored red. The pink shaded regions denote NAB's anomaly windows;
any detection within a window is considered a true positive.}
\label{results}
\end{center}
\vskip -0.2in
\end{figure*}

We ran our HTM based anomaly detector on NAB.  Table~1 contains our scores as
well as scores for several other algorithms, including Etsy Skyline, Twitter
AnomalyDetectionVec (ADVec), and a variant of Bayesian online change point
detection \cite{Adams2007}\footnote{Please see the NAB repository at
\url{https://github.com/numenta/NAB} for specific implementation and parameter
tuning details.}. We include the score for a ``Perfect'' detector, i.e. an
idealized detector that detects every anomaly as early as possible and generates
no false positives. We also include the scores for ``sliding threshold'' and
``random'' detectors as baselines for comparison. The HTM detector achieves the
best overall score, followed by Twitter ADVec and Etsy Skyline.
However, as can
be seen by the performance of the Perfect detector, the benchmark is a
challenging one with significant room for future improvements.

The NAB benchmark doesn't measure an explicit ROC curve but includes two
additional ``application profiles'' that tradeoff the weighting between false
positives and false negatives. The ``reward low FP'' profile applies a higher cost
to false positives. Conversely the ``reward low FN'' profile applies a higher cost
to false negatives. The last two columns of Table~1 show our
scores for those two profiles. Although there are individual variations (Twitter
in particular improves significantly in the ``reward low FN'' column), our
detector performs best overall indicating that it achieves a good tradeoff
between false positives and false negatives.

A closer look at the errors illustrates some interesting situations
that arise in real time applications.
Figure~\ref{results}(a) demonstrates the value of continuous learning. This file
shows CPU usage on a production server over time and contains two anomalies. The
first is a simple spike detected by all algorithms. The second is a sustained
shift in the usage. Skyline and HTM both detect the change but then adapt to the
new normal (with Skyline adapting quickest). Twitter ADVec however continues to
generate anomalies for several days.  Skyline scored highest on this stream.

Figure~\ref{results}(b) and (c) demonstrate temporal anomalies and their
importance in early detection. Figure~\ref{results}(b) is an example
demonstrating early detection. All three detectors detect the anomaly but HTM
detects it three hours earlier due to a subtle shift in metric dynamics.
Figure~\ref{results}(c) shows a close up of the results for the machine temperature
sensor data shown in Figure~\ref{icml-historical}. The anomaly on the left is a
somewhat subtle temporal anomaly where the temporal behavior is unusual but
individual readings are within the expected range. This anomaly (which preceded
the catastrophic failure on February 8) is only detected by HTM.  All three
detectors detect the anomaly on the right, although Skyline and HTM detect it
earlier than ADVec. In this plot HTM and Skyline also each have a false
positive.

We chose these examples because they illustrate common situations that occur in
practice. Qualitatively we have found that temporal changes in behavior often
precede a larger, easily detectable, shift. Temporal and sequence based anomaly
detection techniques may be able to detect anomalies in streaming data before
they are easily visible. We speculate that the early detection in
Figure~\ref{results}(b) is due to the temporal modeling in HTMs as the earlier
shift is difficult to detect through purely spatial means. This provides hope
that such algorithms can be used in production to provide early warning and
perhaps help avoid problems far more reliably than spatial techniques.

Low scores for Bayesian change point detection are due to its strong
assumptions regarding a Gaussian or Gamma distribution of data points.
Real-world streaming data is messy and these assumptions do not hold in many
applications. In data where it does hold (e.g. Figs 2, 9a) the algorithm works
well but for most streams (e.g. Figs 1, 4) it does not. Indeed, it is this lack
of any consistent distribution across streaming applications that led us to
model the distribution of anomaly scores rather than the distribution of metric
values.

%MENTION THAT NAB ONLY USES SINGLE MODELS

From the standpoint of computational efficiency, on the first author's laptop,
it takes 48 minutes to run NAB's complete dataset of 365,558 records. This
represents an average of 8 milliseconds per record.

\section{Discussion}

With the increase in connected real-time sensors, the detection of anomalies in
streaming data is becoming increasingly important. The use cases cut across a
large number of industries; anomaly detection might represent the most
significant near-term application for machine learning in IoT.

In this paper we have discussed a novel anomaly detection algorithm for real-time
streaming applications. Based on HTMs, the algorithm is capable of detecting
spatial and temporal anomalies in predictable and noisy domains. A probabilistic
formulation allows the user to control the rate of false positives, an important
consideration in many applications. We have discussed an extension to large
systems with multiple independent models that incorporate temporal windows.

Our results show that the algorithm can achieve best in class results on a
benchmark of real world data sources. Our system is practical in that it is
computationally efficient, automatically adapts to changing statistics, and
requires little to no parameter tuning. It is currently in use in commercial
applications, and the complete source code to the algorithm is available as open
source software\footnote{Please see \url{http://numenta.com/nab}}.

There are a number of possible extensions to the algorithm. The error
analysis from NAB indicates that the errors from our algorithm are not
necessarily correlated with errors from the other two algorithms. As such an
ensemble based approach might provide a significant increase in accuracy.
The assumption of a Gaussian distribution for anomaly scores is not
always correct. Exploring other distributions represents another possible
extension that could potentially improve the results.

\section*{Acknowledgements} 
 
We thank Alex Lavin, Yuwei Cui, and Jeff Hawkins for many helpful comments
and suggestions.

\bibliography{htm_bibliography}
\bibliographystyle{icml2016}

\end{document}